\documentclass[conference]{IEEEtran}
\IEEEoverridecommandlockouts
\usepackage{cite}
\usepackage{amsmath,amssymb,amsfonts}
\usepackage{array}
\usepackage{algorithmic}
\usepackage{graphicx}
\usepackage{textcomp}
\usepackage{xcolor}
\usepackage{makecell}
\usepackage{subcaption}
\usepackage{threeparttable}
\usepackage{booktabs}
\usepackage{fontawesome}
\usepackage{float}
\usepackage{enumitem}
\def\BibTeX{{\rm B\kern-.05em{\sc i\kern-.025em b}\kern-.08em
    T\kern-.1667em\lower.7ex\hbox{E}\kern-.125emX}}

\begin{document}

\title{
    Dynamic Neural Field Modeling of Visual Contrast for Perceiving Incoherent Looming
    }

% \thanks{This research has been funded by the National Natural Science foundation of China under the Grant No.12031003, No. 12211540710, No. 62376063, the Social Science fund of the Ministry of Education of China under the Grant No. 22YJCZH032, the Guangzhou University Graduate Student Overseas Joint Training Program and the Innovation Research Grant for the Postgraduates of Guangzhou University.}

%         <-this % stops a space

% \author{\IEEEauthorblockN{Ziyan Qin}
% \IEEEauthorblockA{\textit{School of Mathematics and Information Science} \\
% \textit{Machine Life and Intelligence Research Centre}\\
% \textit{Guangzhou University}\\
% China \\
% \textit{School of Computing and Mathematical Sciences} \\
% \textit{University of Leicester}\\
% United Kingdom \\
% ziyan9603@e.gzhu.edu.cn}
% \and
% \IEEEauthorblockN{Qinbing Fu}
% \IEEEauthorblockA{\textit{School of Mathematics and Information Science} \\
% \textit{Machine Life and Intelligence Research Centre}\\
% \textit{Guangzhou University}\\
% China \\
% qifu@gzhu.edu.cn}

% \and
% \IEEEauthorblockN{Jigen Peng}
% \IEEEauthorblockA{\textit{School of Mathematics and Information Science} \\
% \textit{Machine Life and Intelligence Research Centre}\\
% \textit{Guangzhou University}\\
% China \\
% jgpeng@gzhu.edu.cn}

% \and
% \IEEEauthorblockN{Shigang Yue}
% \IEEEauthorblockA{\textit{School of Computing and Mathematical Sciences} \\
% \textit{University of Leicester}\\
% United Kingdom \\
% sy237@le.ac.uk}
% }

 \author{
 \IEEEauthorblockN{Ziyan Qin\textsuperscript{1,2}, Qinbing Fu\textsuperscript{*,1}, Jigen Peng\textsuperscript{1}, Shigang Yue\textsuperscript{2}}\\
 \IEEEauthorblockA{
 \textsuperscript{1}\textit{Machine Life and Intelligence Research Center, School of Mathematics and Information Science, Guangzhou University, China}\\
 \textsuperscript{2}\textit{School of Computing and Mathematical Sciences, University of Leicester, United Kingdom}
% Emails: ziyan9603@e.gzhu.edu.cn, qifu@gzhu.edu.cn, jgpeng@gzhu.edu.cn, sy237@le.ac.uk
 }
 }
%\author{
%\IEEEauthorblockN{Anonymous Authors}}

\maketitle

\begin{abstract}
% The Amari's Dynamic Neural Field (DNF) framework, which models the average activation of neuronal groups, has emerged as a promising foundation for developing low-energy looming perception methods in robotic applications. While these single-field approaches are efficient, they face significant challenges in detecting incoherent stimuli—conditions frequently encountered in real-world environments, such as during rain.
Amari's Dynamic Neural Field (DNF) framework provides a brain-inspired approach to modeling the average activation of neuronal groups. Leveraging a single field, DNF has become a promising foundation for low-energy looming perception module in robotic applications. However, the previous DNF methods face significant challenges in detecting incoherent or inconsistent looming features—conditions commonly encountered in real-world scenarios, such as collision detection in rainy weather. Insights from the visual systems of fruit flies and locusts reveal encoding ON/OFF visual contrast plays a critical role in enhancing looming selectivity. Additionally, lateral excitation mechanism potentially refines the responses of loom-sensitive neurons to both coherent and incoherent stimuli. Together, these offer valuable guidance for improving looming perception models. Building on these biological evidence, we extend the previous single-field DNF framework by incorporating the modeling of ON/OFF visual contrast, each governed by a dedicated DNF. Lateral excitation within each ON/OFF-contrast field is formulated using a normalized Gaussian kernel, and their outputs are integrated in the Summation field to generate collision alerts. Experimental evaluations show that the proposed model effectively addresses incoherent looming detection challenges and significantly outperforms state-of-the-art locust-inspired models. It demonstrates robust performance across diverse stimuli, including synthetic rain effects, underscoring its potential for reliable looming perception in complex, noisy environments with inconsistent visual cues.

% Looming perception is a fundamental capability for both biological and artificial systems, as it provides one of the simplest yet most effective cues for detecting impending collisions. Despite significant progress in collision detection through dynamic neural field models and insect-inspired visual systems, the challenges posed by incoherent looming stimuli, particularly in complex and noisy environments, remain largely unaddressed. In this paper, we present a novel contrast-based dynamic neural field (C-DNF) model to tackle these limitations. Our approach enhances ON/OFF contrast signals through lateralized excitation, while isolating and suppressing noise via a combined processing field. Experimental results validate the effectiveness of the proposed method, showcasing robust performance on both synthetic incoherent looming stimuli and complex, rainy real-world scenarios.
\end{abstract}

\begin{IEEEkeywords}
dynamic neural field, looming perception, incoherence, ON/OFF visual contrast, lateral excitation
\end{IEEEkeywords}

\section{Introduction}
% para1: dynamic neural field is a brain inspired frame work and shows wide range application on cognitive robotics and learning, recent study have been extended its application to looming perception due to its marvelous ability on processing the continuous video streams.
% para2: DNF for looming perception and the importance of looming perception, and the limitation of the previous single fielded DNF: limiataions in incoherence looming perception, while incoherent looming perception is rather ubiquitous in real world enviroment, such as rainy days.
% para3: We can always get the wanted answer from nature, studies of the visual systems in Drosophila and locusts have demonstrated that ON/OFF visual pathways play a crucial role in enhancing looming selectivity. Additionally, lateral excitation mechanisms have been shown to fine-tune the neuronal responses of looming-sensitive neurons to both coherent and incoherent looming stimuli. models for looming perception have been emerged from one and another, but they have never been tested under incoherence looming stimuli.
% para4: thus, we build up this model, the combinition field is used for denoise while the on/OFF pathway is used for signal enhancement.
% [fIGURE1: model structure and samples for incoherence stimuli and rainy stimuli]
% para5: Contributions:
% 1. the contrast based DNF, and shows great performance on incoherence looming and complex real world environment;
% 2. the SOTA LGMD-based models have been tested and most of their looming selectivity remains under the incoherence motion stimuli.

Inspired by the functional organization of the cortex, Amari's Dynamic Neural Field (DNF) framework, introduced in the 1970s, provides a computational approach to model the average activation dynamics of neuronal groups with similar functionalities \cite{Amari1977,Giese1999}. These activation dynamics are primarily governed by lateral interactions and input stimuli. By coupling multiple interacting DNFs, this computational framework has been successfully applied to model various cognitive processes, including memory and learning \cite{Camperi1998,Schoner2002,Kamkar2022,Qin2022,Wojyak2021}.

Due to their impressive ability to process continuous image streams, recent studies have focused on single-field implementations for addressing simple perception tasks, such as collision detection.
Unlike traditional collision detection methods that depend on specific sensors or large training datasets, DNF-based frameworks detect impending collisions by leveraging critical luminance changes caused by the looming of an object, serving as timely collision indicators.
To improve the practicality of the DNF framework for looming perception, a time-delay mechanism was incorporated into the lateral interactions, reducing computational demands and enabling its deployment in real-world scenarios for compact autonomous robots \cite{Qin202x}.
Building on this, Qin et al. introduced adaptive lateral interactions regulated by input luminance intensity. Combined with a dynamic threshold mechanism, this approach enhances looming selectivity, effectively distinguishing looming stimuli from other motion patterns, such as receding, translating, and grating \cite{Qin2024}.

However, these looming perception methods face challenges in detecting incoherent looming stimuli, where the looming object lacks a continuous edge or shape and is instead composed of unevenly distributed spots. Such stimuli are frequently encountered in real-world scenarios, such as identifying looming objects in rainy conditions.

\begin{figure*}[htp]
	\vspace{-20pt}
    \centering
    \includegraphics[width=0.88\linewidth]{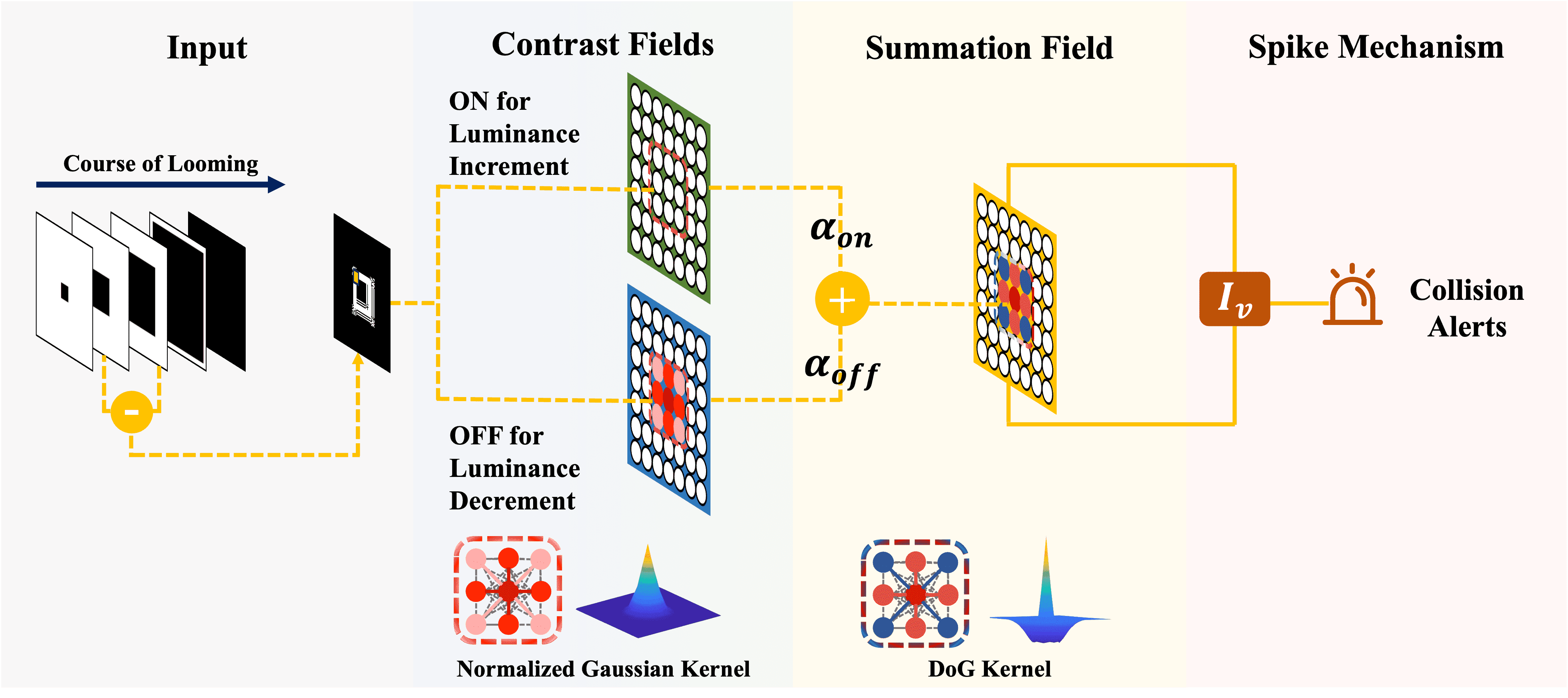}
    \caption{The diagram of the proposed C-DNF model for looming perception. 
    The proposed framework consists of three distinct DNF fields: ON-contrast, OFF-contrast, and Summation field. The model input is the luminance change between successive video frames, with each neuron corresponding to a pixel in the input video (e.g., the yellow pixel in the input corresponds to the red neuron in the OFF field for demonstration). For a resolution of $m\times n$, each field comprises $m\times n$ neurons.
    Luminance increments and decrements are separated serving as input to the respective fields. Neurons within the ON/OFF-contrast fields receive excitatory signals from their $8$ immediate neighbors (as shown by red and pink neurons in the inset), modulated a normalized Gaussian kernel (3D inset).
    In contrast, the lateral interaction within the Summation field is regulated by a Difference of Gaussian (DoG) function (3D inset), where proximal neurons provide excitation (depicted as red neurons in the inset) and distal neurons provide inhibition (depicted as blue neurons in the inset).
    The processed ON and OFF signals, scaled by contrast coefficient $\alpha_{on}$ and $\alpha_{off}$ flow into the Summation field. Neuron activations in the Summation field are integrated into an overall signal $I_v$ via a sigmoid-like function, triggering collision alert when $I_v$ exceeds a predefined threshold.
    Notably, dashed yellow lines in the diagram indicate pixel-wise computations, while solid yellow lines represent integration of the entire neuron field into a single signal.}
    \label{fig:model structure}
    \vspace{-10pt}
\end{figure*}

Nature often holds the solutions we seek. Anatomical studies have uncovered specialized loom-sensitive neurons in insects, such as the lobula giant movement detectors (LGMD1 and LGMD2) in locusts \cite{Oshea1974,Rind1996a} and the lobula plate/lobula columnar type-II (LPLC2) neurons in fruit fly \textit{Drosophila} \cite{Ache2019,Klapoetke2017}. These neurons, with their highly effective afferent networks and precise neuronal computations, are finely tuned for looming perception, enabling insects to navigate efficiently through cluttered groups and complex environments.

Research on LGMDs and LPLC2 has shown the encoding of ON/OFF visual contrast, which process luminance increments (ON-contrast) and decrements (OFF-contrast) separately, are essential for enhancing looming selectivity \cite{fu2023,fu2020}. Lateral excitation mechanisms further amplify signals, enabling loom-sensitive neurons to detect both coherent stimuli with clear edges and incoherent stimuli with fragmented patterns \cite{Rind2016,Zhu2018,Dewell2018,Rind2024}.
Building on these findings, insect-inspired looming perception models have evolved, delivering low-energy, efficient, real-time detection methods.
Fu et al. developed a model simulating the LGMD1 neuron’s looming selectivity by incorporating ON/OFF channels and spike frequency adaptation mechanism. This model demonstrates strong responses to looming stimuli in both contrasts and facilitates efficient collision detection in micro-mobile autonomous robots \cite{fu2018}.
In another study, Fu et al. replicated the LGMD2 neuron’s specific looming selectivity using adaptive inhibition mechanisms with multi-scale time delays, enabling it to respond exclusively to OFF-contrast approaching objects \cite{fu2020}.
Recent advancements have further improved these models by incorporating lateral excitation mechanisms early in the signal processing and combining the looming selectivity of both LGMD1 and LGMD2 into a unified model. This hybrid model dynamically switches between the two neuron types using an adjustable feedback coefficient \cite{Chang2023}.
However, despite the inclusion of ON/OFF channels and lateral excitation mechanisms, none of these models have been tested on incoherent looming stimuli or evaluated in complex real-world conditions with inconsistent motion features, such as scenarios with rain effects.

To address the outlined challenges, we extend single-field DNF looming perception models by encoding ON/OFF visual contrast, referred to as C-DNF (see Fig. \ref{fig:model structure}), where different contrast signals are processed by two separate fields. Instead of using Difference of Gaussian (DoG) kernels for lateral interactions like previous DNF models, C-DNF applies normalized Gaussian kernels to enhance lateral excitation in each contrast field. The processed ON and OFF signals are then integrated in the Summation field, with lateral interactions once again governed by a DoG kernel.
To validate the efficiency and effectiveness of the proposed method, we conducted systematic experiments, including tests on synthetic incoherent stimuli and real-world scenarios augmented with synthetic rain effects. The results clearly demonstrate the robustness of the C-DNF framework, outperforming several state-of-the-art LGMD-based looming perception methods.
The main achievements of this research are summarized as follows:
\begin{itemize}[leftmargin=*]
    \item This work represents the first encoding of ON/OFF visual contrast in a DNF-based looming perception framework. Combined with a normalized Gaussian kernel for lateral excitation, the proposed C-DNF demonstrated robust performance in detecting incoherent looming stimuli across both synthetic and real-world scenarios, showcasing its potential to address looming perception challenges in complex, noisy environments.
    \item Comparative evaluations with state-of-the-art LGMD-based methods confirmed their retained looming selectivity under incoherent stimuli, validating the efficiency of brain-inspired architectures and underscoring the advancements achieved by the proposed C-DNF model.
\end{itemize}

The remainder of this paper is organized as follows: Section \ref{Sec:formulation} introduces the method formulation and outlines the basic selectivity of the proposed C-DNF. Section \ref{Sec:Exp} presents a comprehensive evaluation, including comparative experiments on synthetic incoherent stimuli, real-world scenarios. Finally, Section \ref{Sec:Dis} discusses the findings and concludes the study.

\section{Method formulation and Basic Selectivity}
\label{Sec:formulation}
This section introduces the formulation of the proposed C-DNF model and its basic selectivity on motion stimuli. Similar to previous single-field DNF methods, the model aligns each field’s receptive area with the resolution of the input video and uses luminance changes as input \cite{Qin202x,Qin2024}. However, unlike earlier approaches, it separates luminance changes into ON/OFF visual contrasts, which independently drive the neuronal activation of their respective fields. Neurons in the Summation field integrate these dynamics uniformly and generate spikes based on a fixed threshold, serving as collision alerts.

The C-DNF computes the stationary solution of each field for every video frame. It exhibits strong responses to looming stimuli, symmetric responses to receding stimuli, and no response to translating.
% To adapt Amari's DNF framework to image streams, the receptive field of the DNF is set to match the resolution of each image in the input stream. After solving the stationary solutions for the ON and OFF DNFs, the resulting signals are integrated in a Summation field. The activation of neurons within the Summation field is uniformly contributed to generating spikes based on a fixed threshold, ultimately producing collision alerts.lateral excitation is incorporated into both ON and OFF contrasts by replacing the standard lateral interaction DoG kernel with a normalized Gaussian kernel. The processed ON and OFF signals are then equally integrated in the summation field to produce the final model response. The proposed model exhibits a strong response to looming stimuli, symmetric responses to receding stimuli, and remains inactive for translating and elongation stimuli.
% The proposed model take the resolution of the video as the receptive field of each DNF fields, the pixels in the video are corresponds to the neurons within each field one to one. The proposed C-DNF compute the stationary solution of each neurons within the fields frame to frame. , with the input stimuli for each neuron corresponding to the luminance change of the associated pixel. Expanding the DNF framework with ON/OFF contrast, the luminance changes are first separated into ON and OFF channels, serving as input stimuli for the respective ON and OFF DNFs.

\subsection{C-DNF Model for Looming Perception}
Assume the video resolution is $m\times n$, then the ON-contrast, OFF-contrast, and Summation fields each consist of $m\times n$ neurons, with a one-to-one correspondence between neurons and video pixels across the three fields. The activation dynamics of these neurons are governed by Amari's standard DNF equation, differing in their input stimuli and lateral interaction kernels. Luminance increments and decrements are taken as input and processed in the ON and OFF fields, respectively. The Summation field integrates the processed signals from both fields to produce a unified response for looming perception.

Let $L(x,y,t) \in \mathbb{R}^3$ denote the input video, where $x,y$ and $t$ denote spatial and temporal coordinates. The luminance change between successive frames is defined as:
$$P(x,y,t) = L(x,y,t)-L(x,y,t-1),$$ 
where $L(x,y,t)$ and $L(x,y,t-1)$ are the gray-scaled brightness of two successive frames, normalized to the range $[0,1]$. 
The luminance changes are separated into ON and OFF channels using half-wave rectification, defined as:
\begin{equation}
\label{eq:P_onOFF}
\begin{aligned}
P_{on}(x,y,t) &= [P(x,y,t)]^+  \\
P_{off}(x,y,t) &= -[P(x,y,t)]^-, \\
\end{aligned}
\end{equation}
where $[x]+$ and $[x]-$ denote $max(x,0)$ and $min(x,0)$ respectively. Here, $P_{on}(x,y,t)$ captures the luminance increment as ON contrast, while $P_{off}(x,y,t)$ captures the luminance decrement as OFF contrast.
The $P_{on}(x,y,t)$ and $P_{off}(x,y,t)$ flow into the ON/OFF-contrast fields and serves as their input stimuli.
The activation behavior of the neurons in ON-contrast fields, $u_{on}(x,y,t)$, can be regulated as 
\begin{equation}
\label{eq:u_on}
\begin{aligned}
    \frac{\partial u_{on}(x,y,t)}{\partial t} &= -u_{on}(x,y,t) + P_{on}(x,y,t) - h \\
    & + \vartheta(\sum^{1}_{i=-1}\sum^{1}_{j=-1}W_c(i,j)u_{on}(x+i,y+j,t)),
\end{aligned}
\end{equation}
where $h = 0.2$ is the resting level of the neuron. The last term of Eq. \ref{eq:u_on} represents the lateral interaction that this neuron receive. Generally, the lateral interaction is regulated by the DoG kernel, which has a 'Mexican Hat' shape, indicating that the neuron receive lateral excitation from proximal neurons and lateral inhibition from distal neurons. However, in the C-DNF, the DoG kernel is replaced by the normalized Gaussian kernel $W_c$, transforming the lateral interaction into lateral excitation. By further narrowing the lateral interaction range to include only the $8$ immediate neighboring neurons, $W_c$ effectively implements lateral excitation. The formulation of $W_c$ is given as:
\begin{equation}
\label{eq:W_c}
W_c(i,j) = \frac{exp(-\frac{i^2 + j^2}{2 \sigma^2_c})}{\sum^{1}_{i=-1}\sum^{1}_{j=-1}exp(-\frac{i^2 + j^2}{2 \sigma^2_c})},
\end{equation}
where $i,j \in \{-1,0,1\}$ represent the neighboring neurons relative to the central neuron, and $\sigma_c = 1$ controls the spatial spread of the excitation.
The threshold function $\vartheta(\cdot)$ for the lateral excitation a neuron receive is defined as:
\begin{equation*}
   \vartheta(u)= \frac{2}{1+e^{-u}} - 1,
\end{equation*}
which also serve as the threshold function of the lateral interaction in Summation field. This function scale the lateral excitation or lateral interaction received by the neuron in to a moderate range of $[-1,1].$

Similarly, the activation behavior of the neurons in OFF-contrast field, $u_{off}(x,y,t)$, can be regulated as 
\begin{equation}
\label{eq:u_OFF}
\begin{aligned}
    \frac{\partial u_{off}(x,y,t)}{\partial t} &= -u_{off}(x,y,t) + P_{off}(x,y,t) - h \\
    & + \vartheta(\sum^{1}_{i=-1}\sum^{1}_{j=-1}W_c(i,j)u_{off}(x+i,y+j,t)),
\end{aligned}
\end{equation}
where $P_{off}(x,y,t)$ is the input stimuli from OFF contrast.

After computing the stationary solutions of Eq. \ref{eq:u_on} and Eq. \ref{eq:u_OFF}, the activated neurons in the ON/OFF-contrast fields are passed as inputs to the Summation field. The activation function for these neurons is defined as:
 $$\theta(u) = \frac{e^u - e^{-u}}{e^u + e^{-u}} \cdot \frac{e^2 + 1}{e^2 - 1}.$$
Notably, this function also serves as the activation function for neurons in the Summation field, maintaining uniformity across the computational framework.
The activation behavior of the neurons in Summation field $v_s(x,y,t)$ is given as:
\begin{equation}
\label{eq:v_s}
\begin{aligned}
    \frac{\partial v_s(x,y,t)}{\partial t} &= -v_s(x,y,t) \\
    & + \alpha_{on} \cdot \theta( u_{on}(x,y,t))+ \alpha_{off} \cdot \theta( u_{off}(x,y,t)) \\
    & -h + \vartheta(\sum^{m}_{i=1}\sum^{n}_{j= 1}W_s(x-i,y-j)v_s(i,j,t)),
\end{aligned}
\end{equation}
For consistency, the resting level in the Summation field is set to $h = 0.2$, and the contrast coefficients $\alpha_{on}$ and $\alpha_{off}$ are both set to 
$$\alpha_{on} = \alpha_{off} = 0.5,$$ indicating that the ON/OFF-contrast signals are contribute equally to the activation of the neurons in Summation field. 
% Specifically, be adjusting the contrast coefficient, the proposed model can shape the specific looming selectivity of the two LGMD neurons.
Unlike the contrast fields, the Summation field reverts to using the DoG kernel for lateral interaction, defined as:
\begin{align}
\label{eq:DOG}
    W_s(x-i,y-j,t) &= A\exp\big({-\frac{(x-i)^2 + (y-i)^2}{2{\sigma}_1^2}}\big) \\ \nonumber
              & -B\exp\big({-\frac{(x-j)^2 +(y-j)^2}{2{\sigma}_2^2}}\big)
\end{align}
where $\sigma_2 = 3\sigma_1$ \cite{Qin2024}, which induce the coefficient as $A = 3/2$ and $B = 1/2$, creating a "Mexican Hat"-shaped interaction profile. We set the excitatory lateral interaction scale $\sigma_1 = 1/3$ in this paper, making the inhibitory lateral interaction scale $\sigma_2 = 1$. Notably, this kernel effectively reduces the response of neurons that receive isolated noise as input, since such neurons would not benefit from lateral excitation from neighboring neurons but are likely to experience lateral inhibition from distal neurons. 

The activated neurons in Summation field are then evenly contributed to the integrated signal $I_v$ through a non-linear sigmoid function:
\begin{equation}
\label{eq:I_v}
    I_v(t) = \frac{1}{1+exp( - \sum^{m}_{i=1}\sum^{n}_{j= 1} \theta(v_s(x,y,t)) \cdot (mn)^{-1})}.
\end{equation}
The proposed C-DNF model ensures that all activated neurons within the Summation field surpass the intrinsic threshold (resting level $h$) before signal integration. A collision alert is triggered when the integrated signal $I_v(t)$ exceeds the initial value of $I_{v0} = \frac{1}{1+exp(0)} = 0.5$ by a small margin. Thus, the threshold for collision alerts is defined as $I_{thre} = I_{v0} + \epsilon$, where $\epsilon = 0.006$ is a predefined margin in this paper. The spike mechanism is described as follows:
\begin{equation}
  \label{eq:spk}
     \text{\text{spike}(t)} = 
     \begin{cases}
     1, & {\text{if}} \ I_v(t) > I_{thre},\\
     {0,} & {\text{otherwise}}.
     \end{cases}
 \end{equation}

Eqs. (\ref{eq:u_on}), (\ref{eq:u_OFF}), and (\ref{eq:v_s}) are computed frame by frame using a fixed-point iteration method. With only the stationary solution used for signal processing, the evolution time during the iteration is disregarded, ensuring that the time variable $t$ aligns with the timestamps of the input video frames. This formulation demonstrates the compact parameter scale of the proposed C-DNF, enhancing its adaptability to diverse looming scenarios, including complex real-world environments.

\subsection{Basic Selectivity of C-DNF}
\label{sec:basic selectivity}
\begin{figure*}[t!]
	\vspace{-10pt}
    \centering
    \begin{subfigure}[t]{0.3\textwidth}
        \centering
        \includegraphics[width=\textwidth]{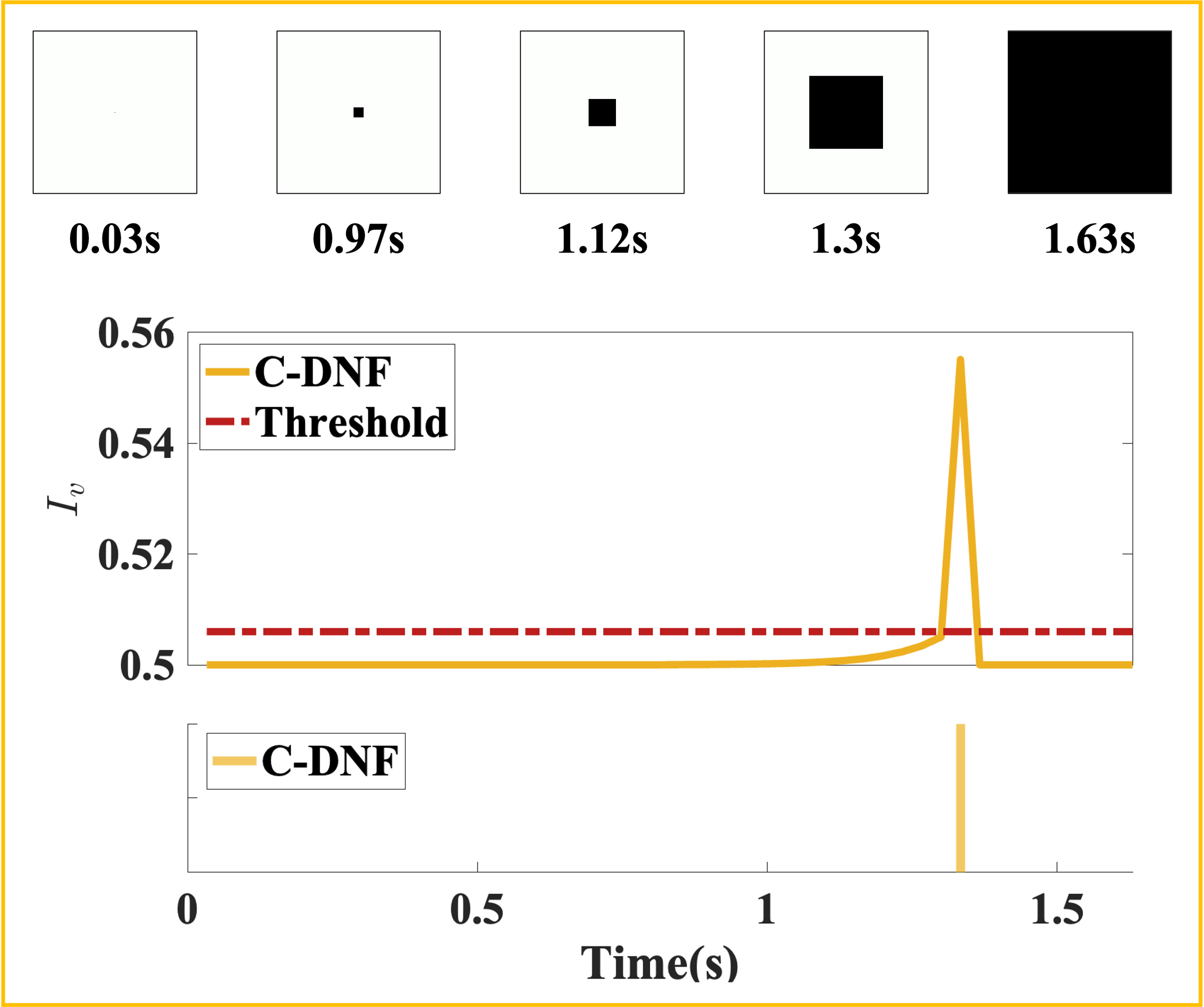}
        \caption{Dark Looming}
        \label{fig:DA}
    \end{subfigure}
    \begin{subfigure}[t]{0.3\textwidth}
        \centering
        \includegraphics[width=\textwidth]{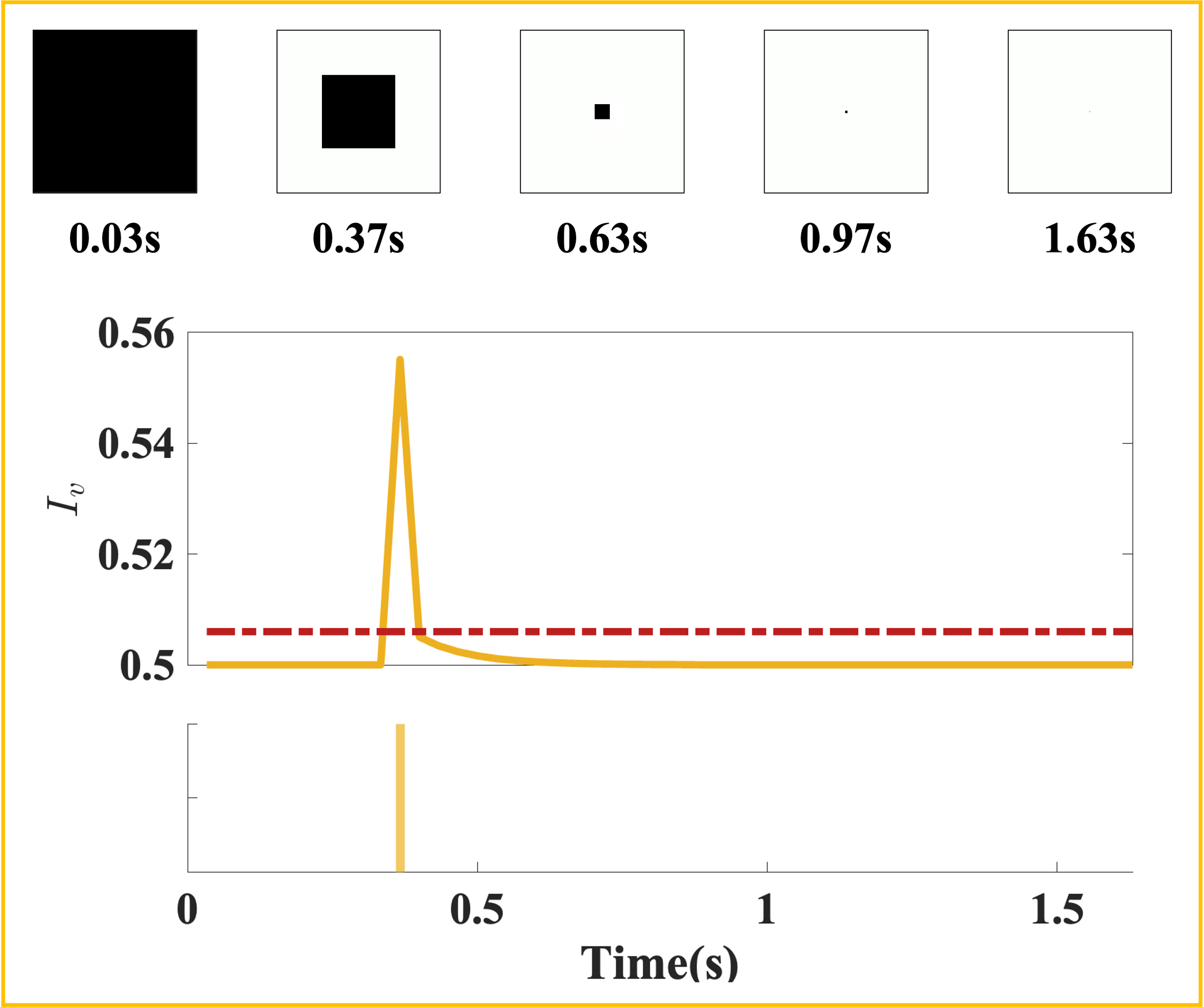}
        \caption{Dark Receding}
        \label{fig:DR}
    \end{subfigure}
    \begin{subfigure}[t]{0.3\textwidth}
        \centering
        \includegraphics[width=\textwidth]{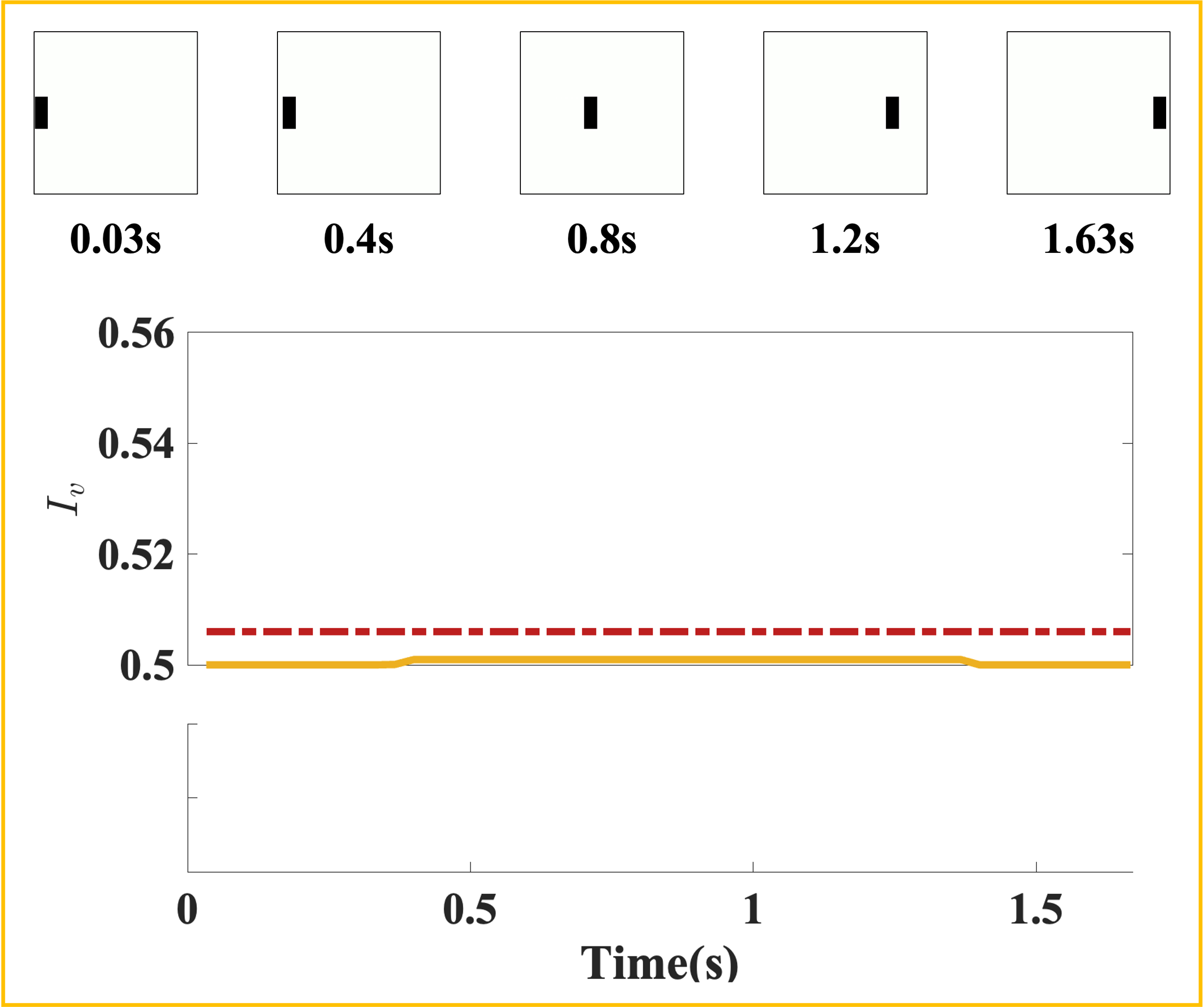}
        \caption{Dark Translating}
        \label{fig:DTH}
    \end{subfigure}

    \begin{subfigure}[t]{0.3\textwidth}
        \centering
        \includegraphics[width=\textwidth]{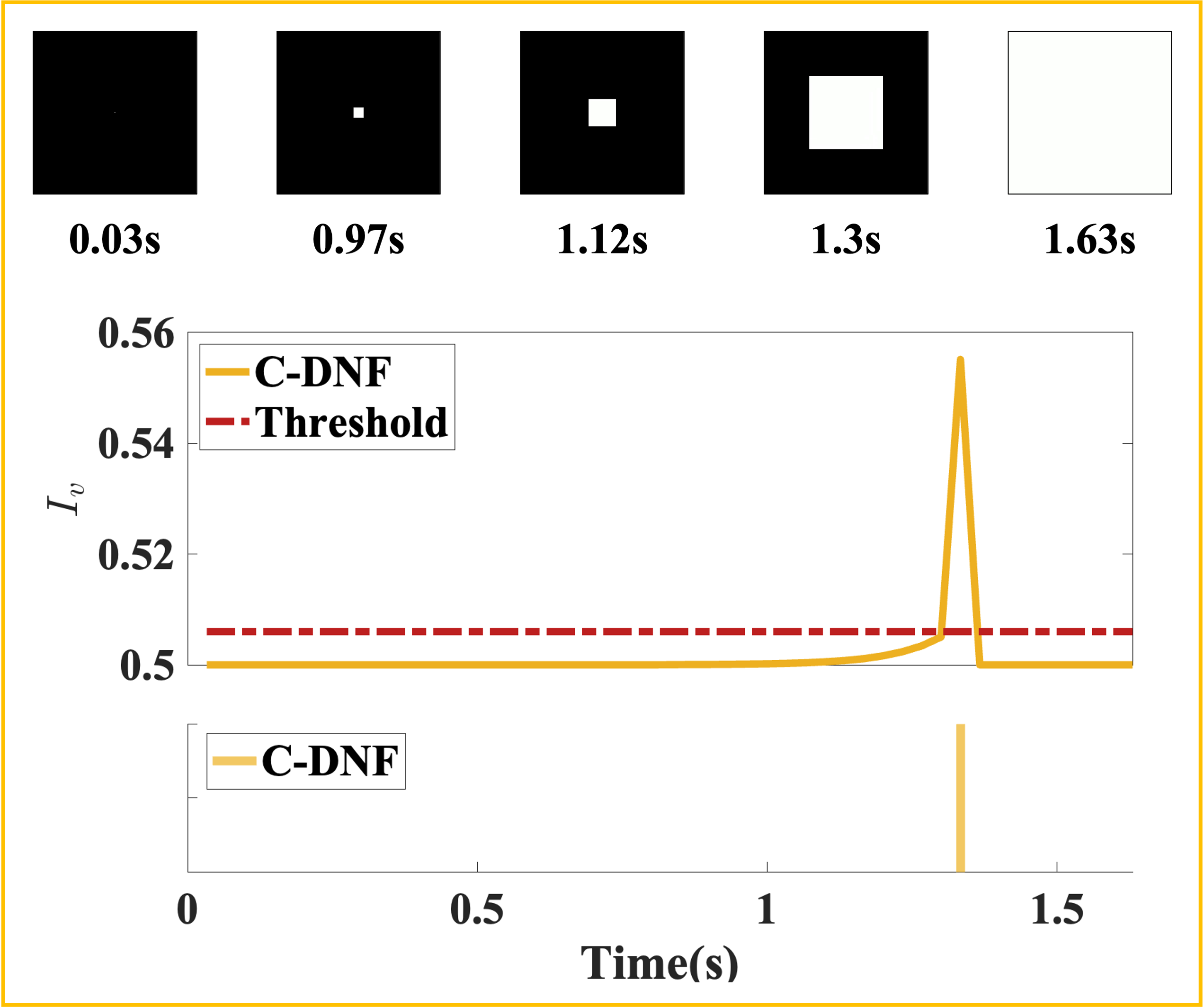}
        \caption{Light Looming}
        \label{fig:LA}
    \end{subfigure}
    \begin{subfigure}[t]{0.3\textwidth}
        \centering
        \includegraphics[width=\textwidth]{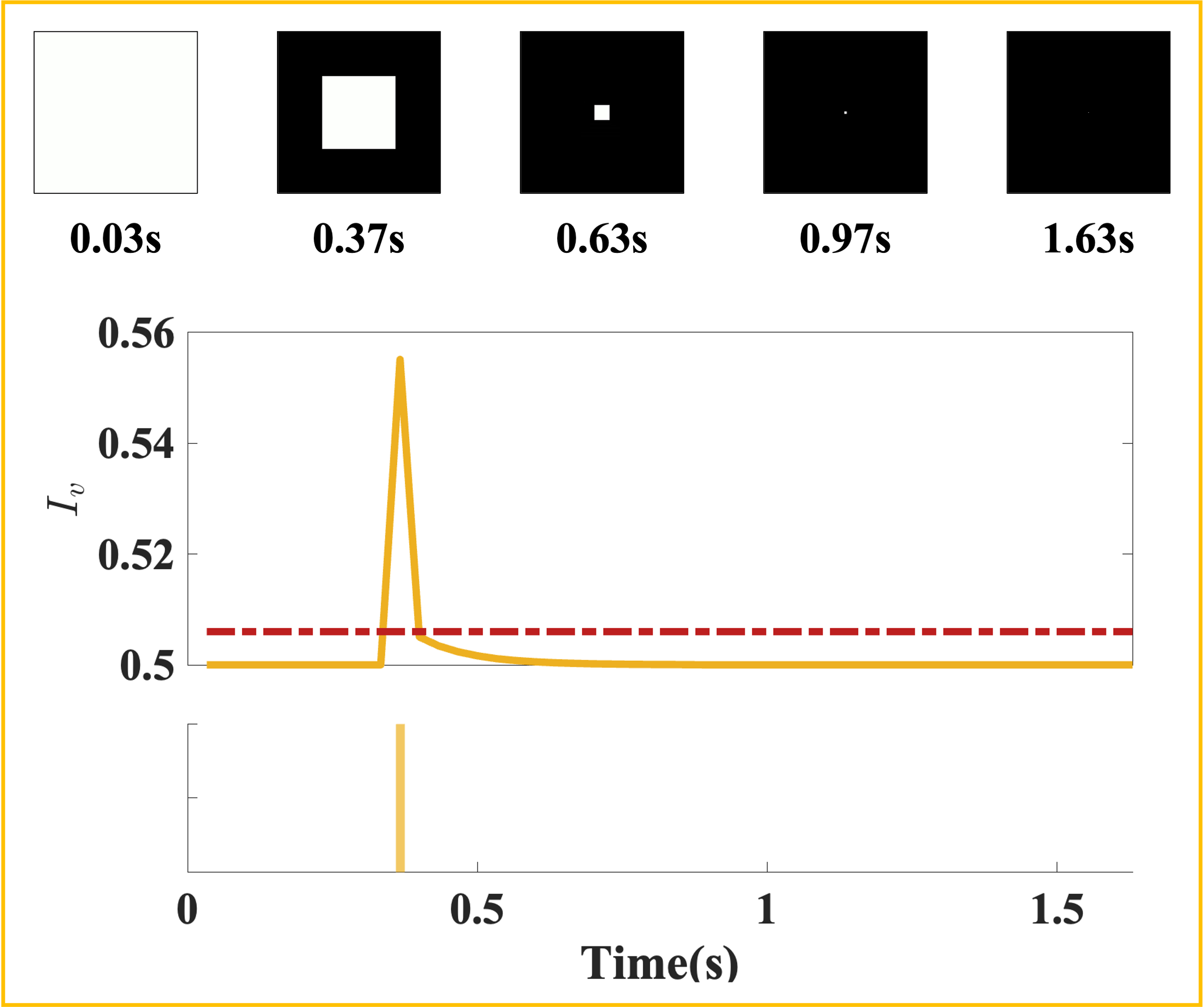}
        \caption{Light Receding}
        \label{fig:LR}
    \end{subfigure}
    \begin{subfigure}[t]{0.3\textwidth}
        \centering
        \includegraphics[width=\textwidth]{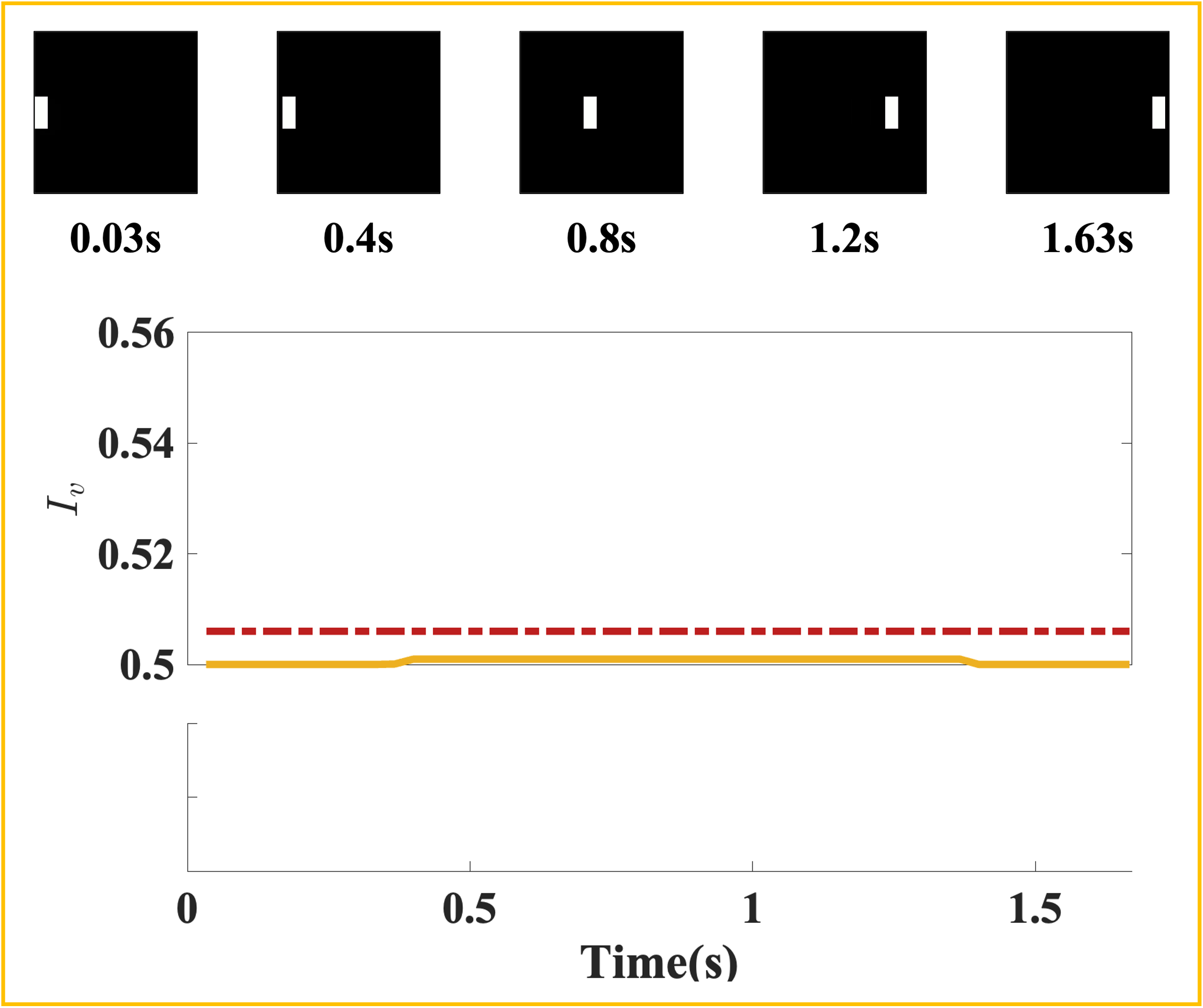}
        \caption{Light Translating}
        \label{fig:LTH}
    \end{subfigure}
    \caption{The model responses of C-DNF to $6$ synthetic stimuli. Each sub-figure comprises three rows: the first row displays snapshots of the tested synthetic stimuli, the second row presents the model's integrated signal $I_v$ along with the fixed threshold $I_{Thre} = 0.506$ and the third row indicates the occurrence of collision alerts. The proposed model demonstrates a robust response to approaching objects irrespective of the contrast, exhibits symmetric responses to receding objects, and no response to translating objects.}
    \label{fig:basic selectivity}
    \vspace{-10pt}
\end{figure*}

The proposed C-DNF model was initially evaluated using simple synthetic stimuli with a resolution $600\times 600 $ pixels at a frame rate of $30$ frames per second. These stimuli included a dark square looming or receding against a white background (see Fig. \ref{fig:DA} and Fig. \ref{fig:DR}), a dark bar translating across the image against a white background (see Fig. \ref{fig:DTH}), and the corresponding looming, receding, and translational motion scenarios with swapped background and object contrasts (see Fig. \ref{fig:LA}, Fig. \ref{fig:LR}, and Fig. \ref{fig:LTH}).

The C-DNF exhibited a strong response to looming objects regardless of contrast and produced symmetrical responses to receding objects, likely due to the reversed nature of the receding stimuli. No collision alerts were generated for translating stimuli, demonstrating the model's selective sensitivity to looming motion (See Fig.\ref{fig:basic selectivity}).

The robustness of the C-DNF was further validated through tests on synthetic incoherent stimuli and recorded real-world stimuli, both with and without synthetic rain effects.

\begin{table}[!t]
%\vspace{-10pt}
\caption{Responses of Various Models to Basic Synthetic Stimuli\label{tab:exp result syn}}
\centering
\begin{threeparttable}
\begin{tabular}{c c c c c}
\hline
Stimuli & C-DNF & SDNF & LGMD2 & F-LGMD ONn \\
\hline
Dark Approaching & 1 (\faCheck) & 1 (\faCheck) & 1 (\faCheck) & 1 (\faCheck) \\
Light Approaching & 1 (\faCheck) & 1 (\faCheck) & 0 (\faCheck) & 1 (\faCheck)\\
\hline
Dark Receding & 1 (\faTimes) & 0 (\faCheck) & 0 (\faCheck) & 1 (\faTimes) \\
Light Receding & 1 (\faTimes) & 0 (\faCheck) & 0 (\faCheck) & 1 (\faTimes)\\
\hline
Dark translating & 0 (\faCheck) & 1 (\faTimes) & 0 (\faCheck) & 0 (\faCheck)\\
Light translating & 0 (\faCheck) & 1 (\faTimes) & 0 (\faCheck) & 0 (\faCheck)\\
\hline
\end{tabular}
\begin{tablenotes}  
        \footnotesize           
        \item[1] $1$ represents that the model generates collision alerts for the corresponding stimulus, while 0 signifies non-response;
        \item[2] (\faCheck) behind the number represents true positive or true negative, while (\faTimes) represents false positive or false negative;
        \item[3] The basic selectivity of the SDNF differs from that reported in the original paper \cite{Qin2024}, which could be attributed to variations in the testing samples or differences in the employed processing systems.
        \end{tablenotes}  
\end{threeparttable}
%\vspace{-10pt}
\end{table}

\section{Experimental Evaluation}
\label{Sec:Exp}
In this section, we evaluate the effectiveness and robustness of the proposed C-DNF model using synthetic incoherent stimuli and a dataset of recorded real-world stimuli, both with and without the synthetic rain effect. The generation processes for these tested stimuli are detailed at the beginning of this section.
For comparison, the C-DNF model is evaluated against the single-field DNF-based method from \cite{Qin2024} (abbreviated as SDNF)and two state-of-the-art insect-inspired models: the LGMD2-based model from \cite{fu2020} (abbreviated as LGMD2) and the LGMD model with negative feedback to ON contrast from \cite{Chang2023} (abbreviated as F-LGMD ONn). The basic selectivity results of the comparative models are presented in Table \ref{tab:exp result syn}.

All experiments in this study, including the basic selectivity tests described in Section \ref{sec:basic selectivity}, were implemented in MATLAB (The MathWorks, Inc., Natick, USA). Data analysis and visualization were also carried out in MATLAB on a MacBook with an M3 Pro chip.
\subsection{Stimuli Generation}
\begin{figure}[htbp]
    \centering
    \begin{subfigure}[t]{0.4\textwidth}
        \centering
        \includegraphics[width=\textwidth]{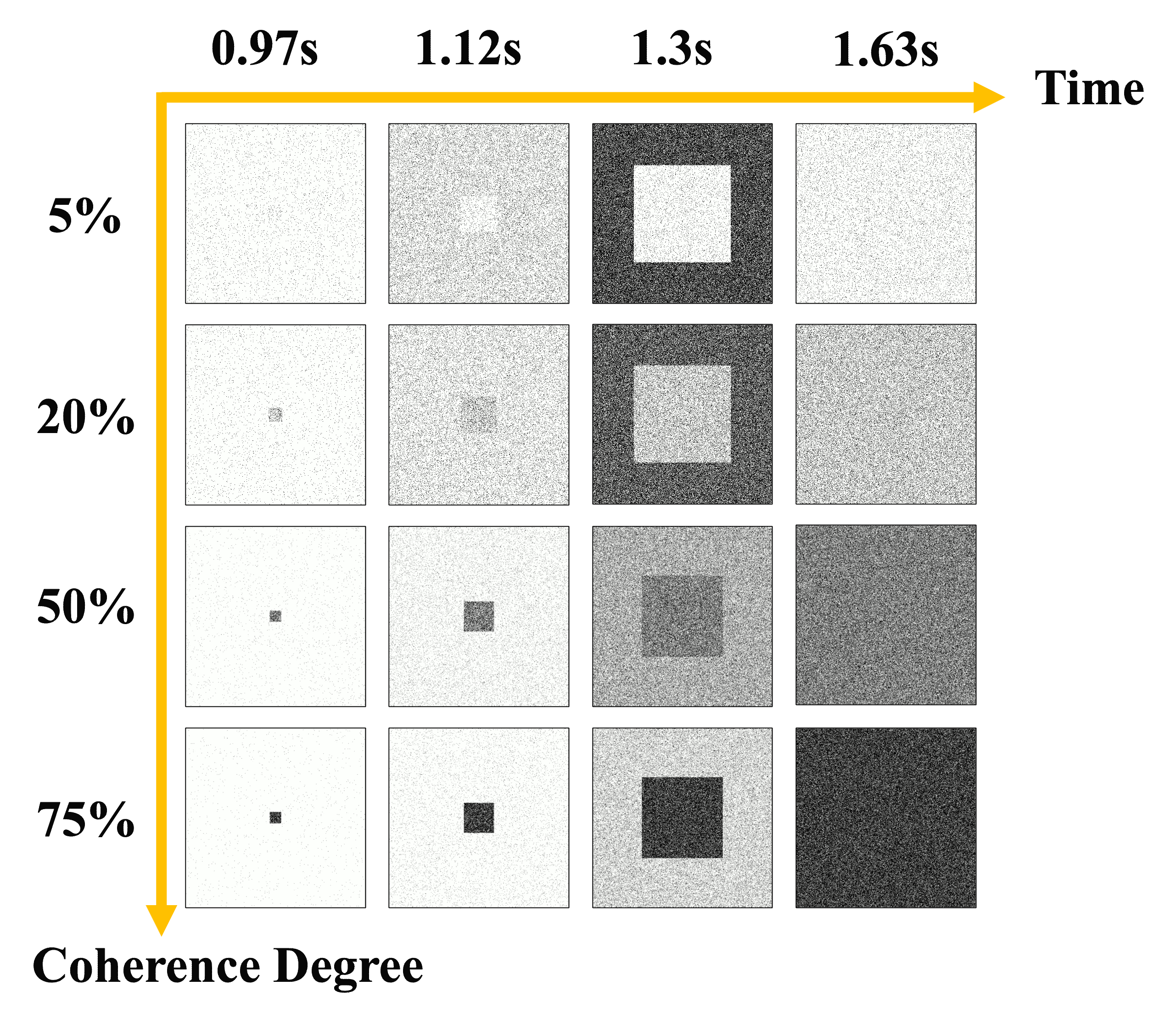}
        \caption{ }
        \label{fig:synthetic}
    \end{subfigure}

    \begin{subfigure}[t]{0.4\textwidth}
        \centering
        \includegraphics[width=\textwidth]{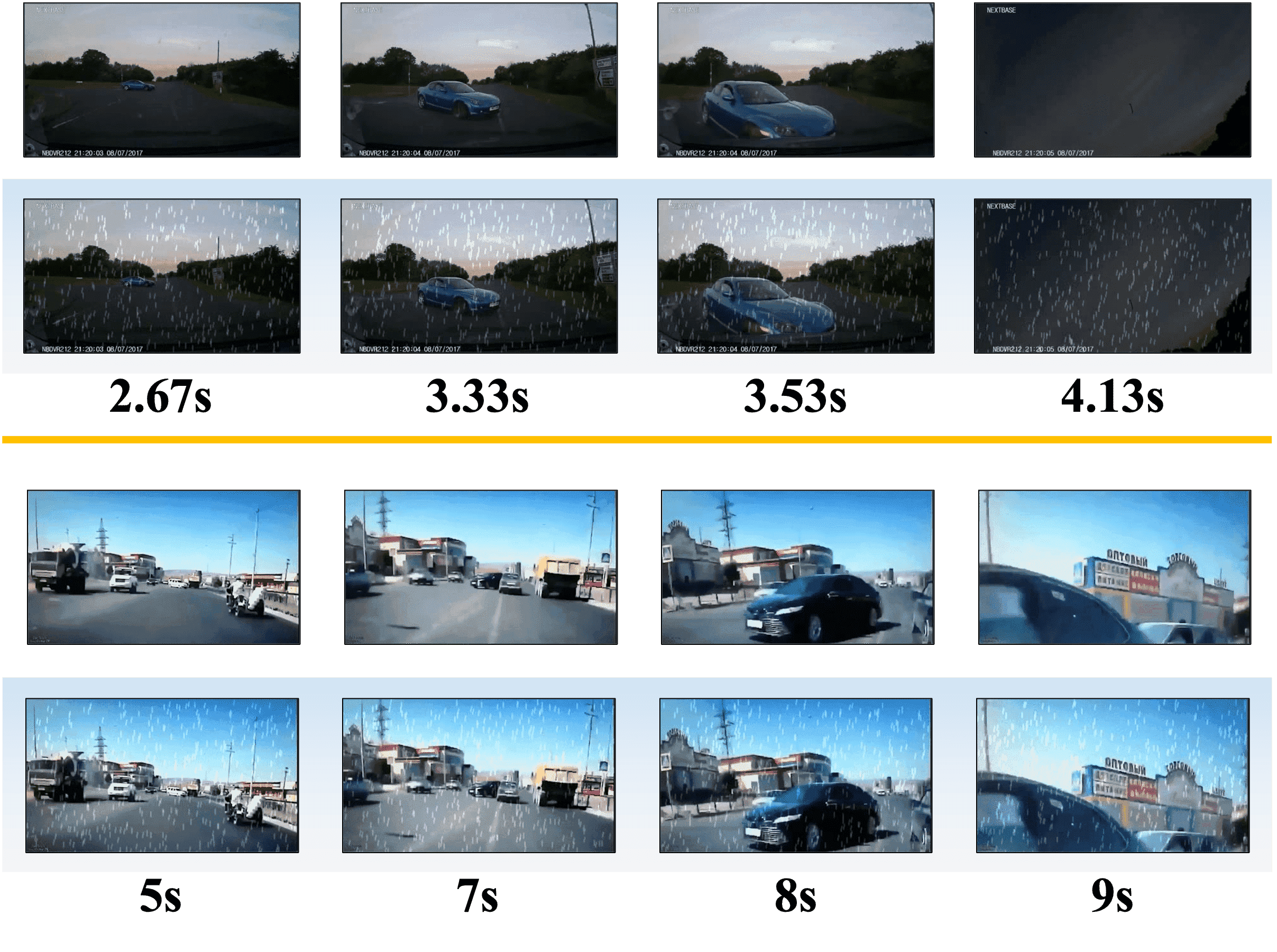}
        \caption{ }
        \label{fig:rain}
    \end{subfigure}

    \caption{Tested samples of synthetic incoherent looming stimuli and real-world stimuli with or without the synthetic rain effect (a) Samples of incoherent looming stimuli with varying coherence levels, including $75\%$, $50\%$, $20\%$ and $5\%$. (b) Two recorded real-world collision scenarios are shown, both with and without synthetic rain effects. The first and third rows display snapshots without rain, while the second and fourth rows show the corresponding scenarios with synthetic rain added.}
    \label{fig:stimuli generation}
    \vspace{-10pt}
\end{figure}

\subsubsection{Synthetic Incoherent Stimuli}
The incoherent stimuli used in this study were generated following a method similar to that described in \cite{Dewell2018}. All $6$ basic synthetic stimuli shown in Fig. \ref{fig:basic selectivity} were utilized to create incoherent stimuli with coherence degrees ranging from $100\%$ to $5\%$. Coherence degree quantifies the continuity and completeness of the edges and shapes of moving objects, such as squares or bars. Stimuli with $100\%$ coherence correspond to standard stimuli (Fig. \ref{fig:basic selectivity}) depicting the regular motion of solid objects, while lower coherence degrees result in less continuous and incomplete shapes, with pixels of the object randomly distributed across the background.
Specifically, the coherence degree is defined as the percentage of pixels that remain in their original positions within the moving object. For example, in Fig. \ref{fig:synthetic}, the $75\%$ coherence dark square looming stimulus depicts a looming process where $75\%$ of the object pixels retain their original positions, as seen in the standard stimuli. The remaining $25\%$ of the pixels are randomly scattered in the background, with their original positions replaced by background pixels. Despite the variations in coherence levels, the incoherent stimuli maintain the same overall luminance change per frame as the standard coherence stimuli ($100\%$ coherence), ensuring consistent physical properties such as resolution and motion speed.
However, when the background cannot adequately accommodate the displaced object pixels, the motion process in the incoherent stimulus halts abruptly, transitioning directly to the final frame where the incoherent moving object attains its maximum size while preserving its incoherent feature.

\subsection{Results on Synthetic Incoherent Stimuli}
\begin{figure*}[htbp]
	\vspace{-20pt}
	\centering
	\begin{subfigure}[t]{0.3\textwidth}
		\centering
		\includegraphics[width=\textwidth]{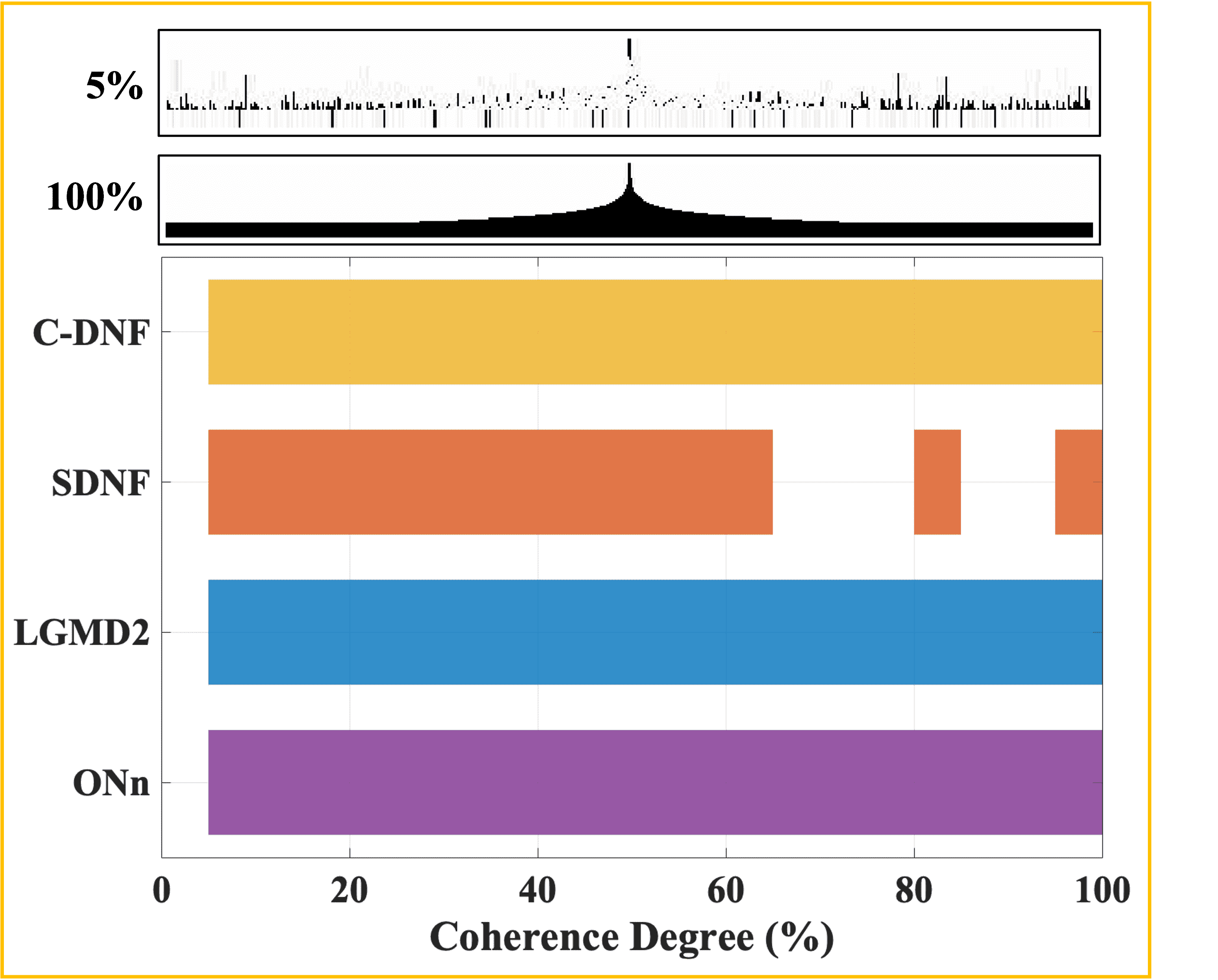}
		\caption{Dark Looming}
		\label{fig:COM EXP CODA}
	\end{subfigure}
	\begin{subfigure}[t]{0.3\textwidth}
		\centering
		\includegraphics[width=\textwidth]{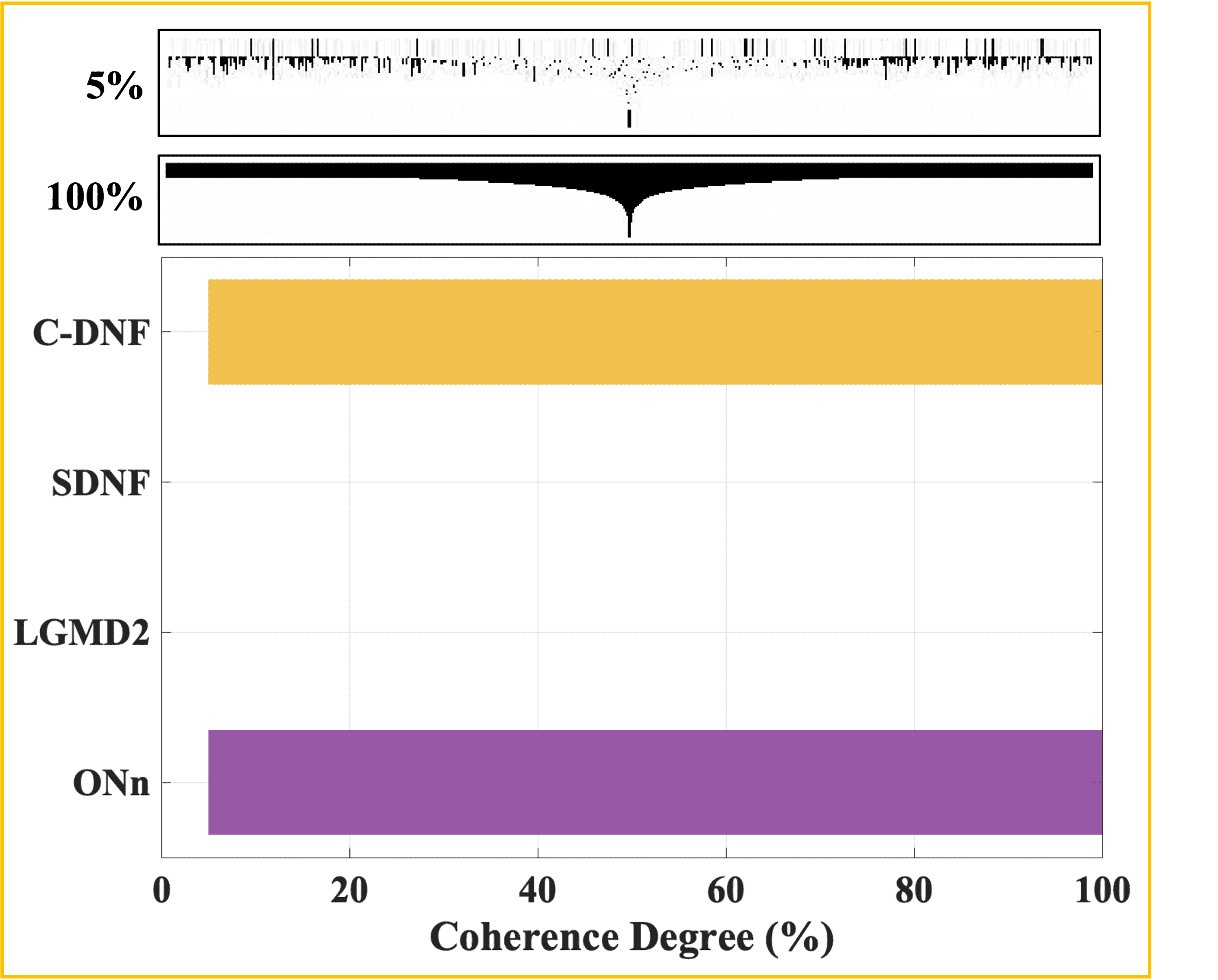}
		\caption{Dark Receding}
		\label{fig:COM EXP CODR}
	\end{subfigure}
	\begin{subfigure}[t]{0.3\textwidth}
		\centering
		\includegraphics[width=\textwidth]{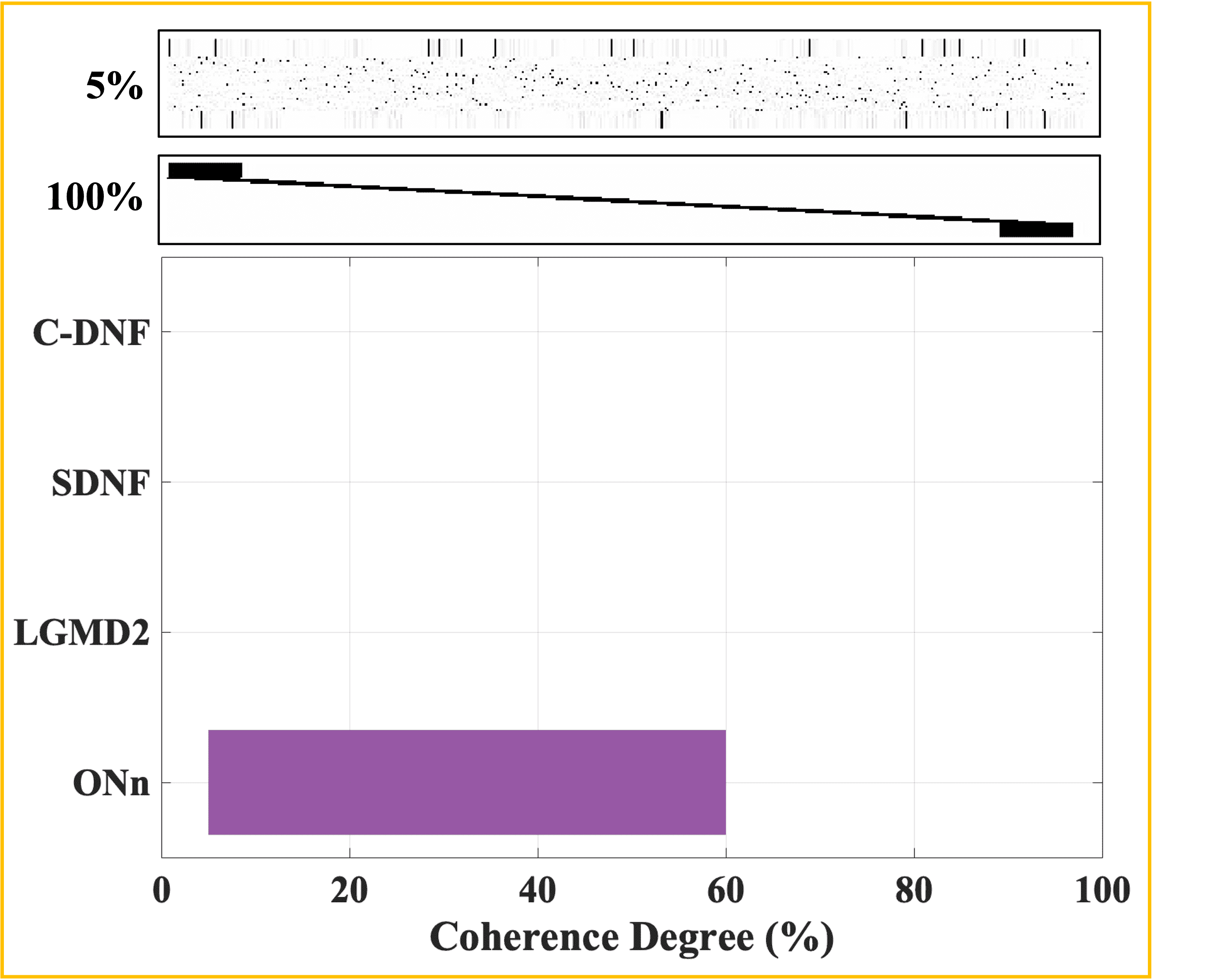}
		\caption{Dark Translating}
		\label{fig:COM EXP CODTH}
	\end{subfigure}
	
	\begin{subfigure}[t]{0.3\textwidth}
		\centering
		\includegraphics[width=\textwidth]{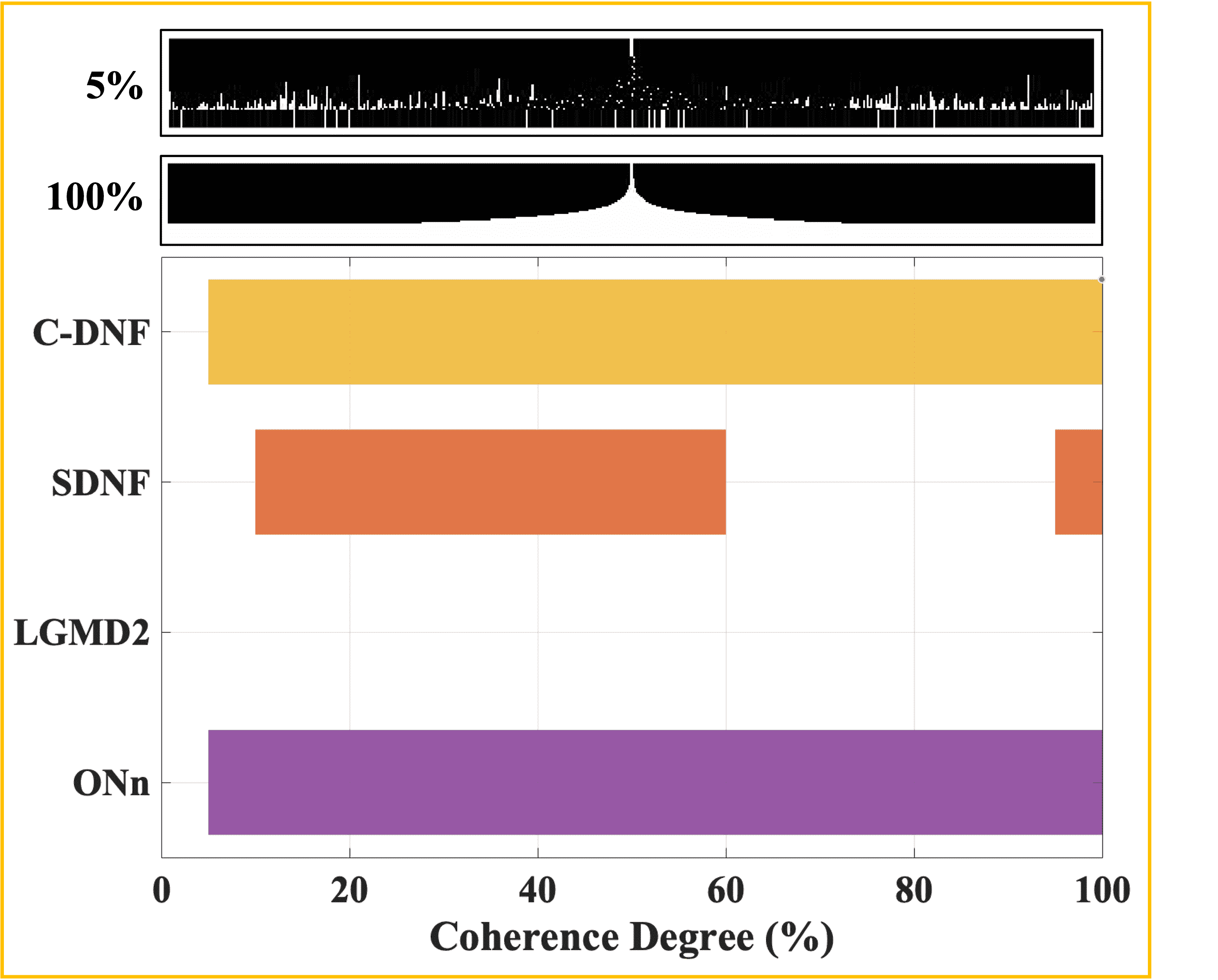}
		\caption{Light Looming}
		\label{fig:COM EXP COLA}
	\end{subfigure}
	\begin{subfigure}[t]{0.3\textwidth}
		\centering
		\includegraphics[width=\textwidth]{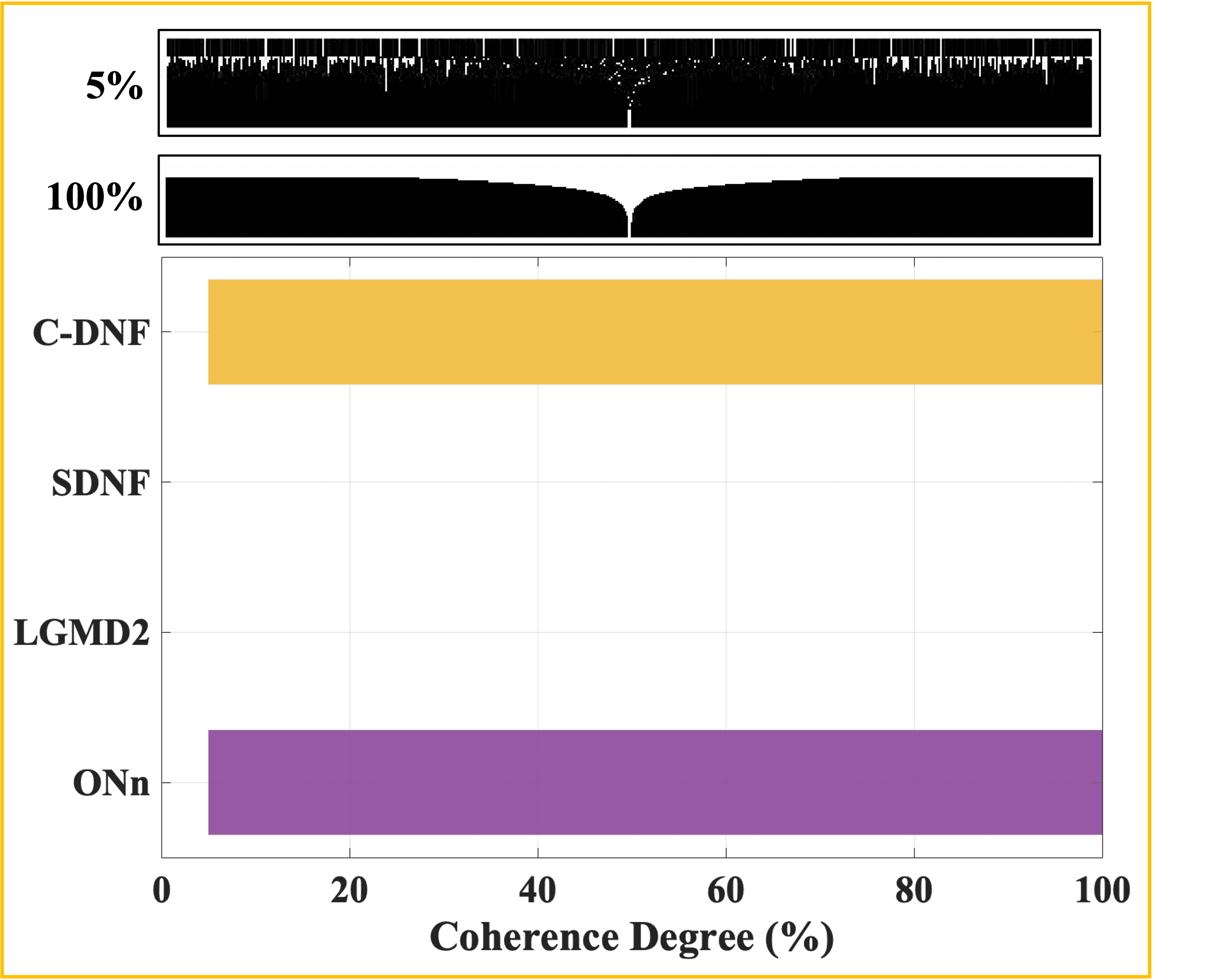}
		\caption{Light Receding}
		\label{fig:COM EXP COLR}
	\end{subfigure}
	\begin{subfigure}[t]{0.3\textwidth}
		\centering
		\includegraphics[width=\textwidth]{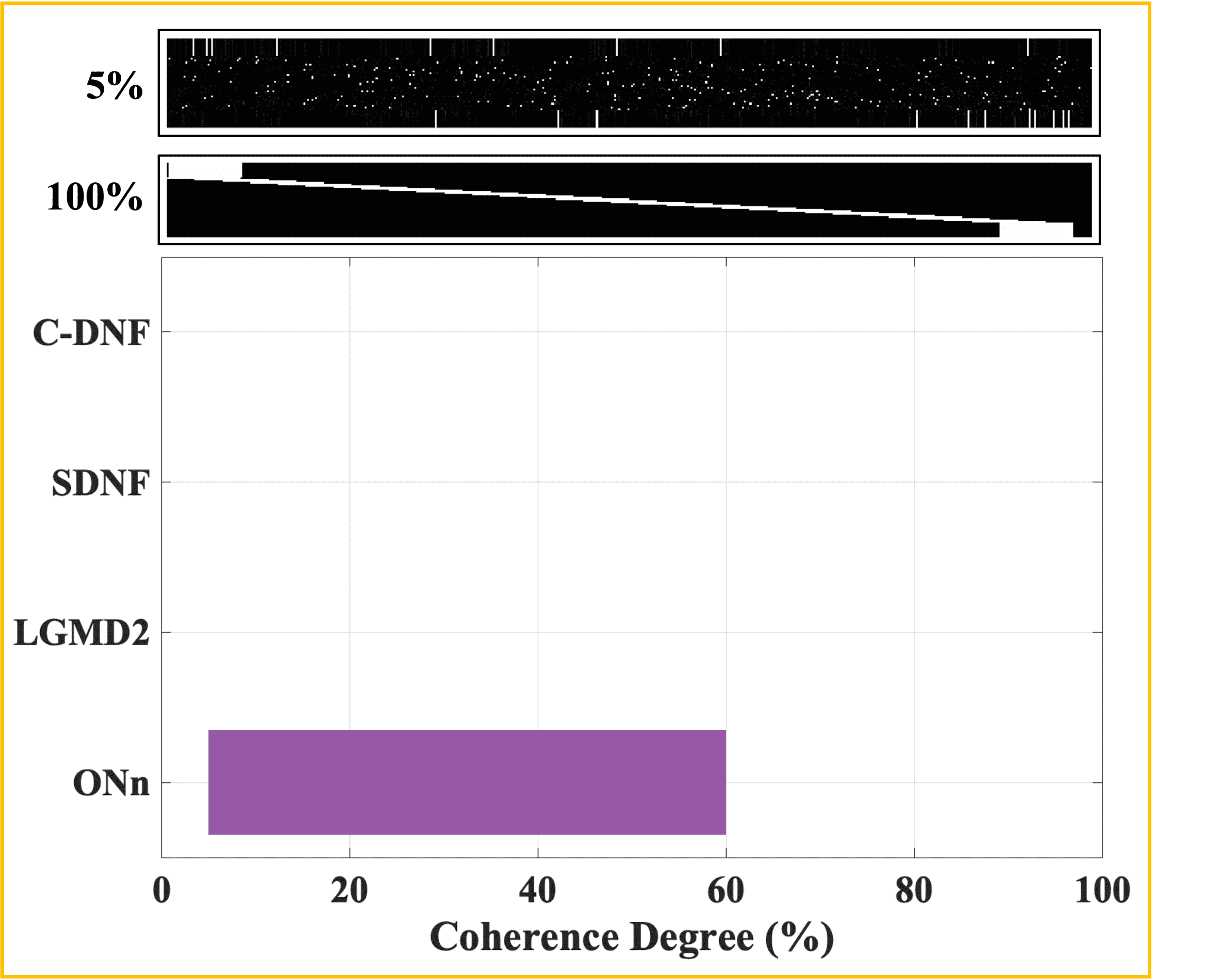}
		\caption{Light Translating}
		\label{fig:COM EXP COLTH}
	\end{subfigure}
	
	\caption{The response characteristics of the comparative models to different motion stimuli with varying coherence degrees. 
		The first and the second rows of each sub-figure illustrate the spatial-temporal moving pattern of the looming, receding and translating object under the coherence degrees of $5\%$ and $100\%$, respectively. The FLGMD ONn is abbreviated as ONn in the figure. The responses of the models are shown as horizontal colored bars beneath the stimulus representation. In general, most of the tested models retained their looming selectivity, as previously reported in Table \ref{tab:exp result syn}, with the exception of FLGMD ONn.}
	\label{fig:coherence exp}
%	\vspace{-10pt}
\end{figure*}

\subsubsection{Recorded Real-world Stimuli with Synthetic Rain Effect}
The recorded real-world stimuli dataset consists of 35 dashboard-recorded road scenarios, of which $20\%$ represent non-collision cases, such as bypassing and translating motions, while the remaining $80\%$ are collision cases. All stimuli have a resolution of $426 \times 240$ pixels with a frame rate of $30$ frames per second. To simulated the incoherence feature in real-world stimuli, a synthetic rain effect was applied to all stimuli, creating a separate dataset.

The synthetic rain effect was generated by overlaying simulated rain droplets onto individual frames to mimic realistic rain conditions. For each frame, a blank layer matching the frame's dimensions was initialized to store the rain effect. A total of 500 droplets were generated per frame, with starting points randomly distributed across the frame. Each droplet extended vertically downward for 8 pixels with a width of 1 pixel and a slight angular deviation to replicate natural rain patterns. The droplets, rendered in gray, were blurred using a Gaussian filter to simulate motion blur.
The rain layer was blended with the original frame through weighted addition, contributing $50\%$ to the final appearance. The spatial distribution of droplets followed a uniform random pattern, with visual streaks suggesting motion. Two examples are presented in Fig. \ref{fig:rain}. The synthetic rain effect introduced noise to the dataset, resulting in average signal-to-noise ratio (SNR) of $19.7 dB$ calculated throughout the whole dataset.

The responses of the comparative models to synthetic incoherent stimuli are summarized in Fig. \ref{fig:coherence exp}. The proposed C-DNF maintained its looming selectivity across all coherence degrees, alongside the LGMD2 and FLGMD ONn models, demonstrating their robustness in detecting incoherent looming stimuli.

In contrast, the SDNF model failed to respond to looming stimuli when the coherence degree ranged from $60\%$ to $95\%$. This is likely because looming stimuli with coherence degrees below $95\%$ were misclassified by the SDNF as noisy looming stimuli due to the excessive scattering of object pixels into the background, which suppressed the model's response. Interestingly, for looming stimuli with coherence degrees below $60\%$, the SDNF model interpreted the stimuli as light objects approaching, thereby preserving its looming selectivity. Additionally, the FLGMD ONn model exhibited responses to translating stimuli when the coherence degree fell below $60\%$. This unexpected behavior may result from the model mistakenly interpreting such stimuli as noisy looming stimuli.

The experimental results on synthetic incoherent stimuli highlight the robustness of the proposed C-DNF model. This robustness is primarily attributed to the lateral excitation mechanism, which amplifies signal intensity in both ON and OFF contrasts, ensuring the preservation of looming selectivity. Similarly, the FLGMD ONn model likely retains its looming selectivity due to its use of Gaussian kernels for processing excitation signals. In contrast, the LGMD2 model demonstrates its ability to maintain looming selectivity through its sophisticated temporal-spatial inhibition mechanisms. These mechanisms, inherent to its elegant model structure, enable the LGMD2 model to effectively handle incoherent stimuli and selectively respond to looming objects.

\subsection{Results on Real-World Stimuli}
In this subsection, we present the comparative experimental results on recorded real-world stimuli, both with and without synthetic rain effects. The perception accuracy of the models is defined as the proportion of tested data in which the model generates correct responses. This metric is calculated using the formula:
\begin{equation}
\label{eq.accuracy}
    accuracy = \frac{TP+TN}{TP+TN+FP+FN} \times 100\%
\end{equation}
where $TP$ (true positives) refers to the number of collision scenarios for which the model correctly generates collision alerts. This is determined by the model issuing its initial alert within a manually set potential collision time range.
$FP $(false positives) denotes the number of non-collision scenarios, such as translating or near-miss stimuli, for which the model incorrectly generates collision alerts. 
$TN$ (true negatives) represents the number of non-colliding stimuli for which the model correctly refrains from generating any collision alerts.
While $FN$ (false negatives) accounts for the number of stimuli for which the model fails to detect a collision.

When tested on the recorded dataset containing both collision and non-collision scenarios, the proposed C-DNF model show an accuracy of $77.14\%$, and maintained its response accuracy even when synthetic rain effects were applied. This robustness is attributed to the enhanced signals in both ON and OFF contrasts, which, in combination with the DoG kernel and the resting level in the Summation layer, effectively suppress noise from isolated signals and maintain the model's accuracy.

\begin{figure*}[htbp]
	\vspace{-20pt}
    \centering
    \begin{subfigure}[t]{0.48\textwidth}
        \centering
        \includegraphics[width=\textwidth]{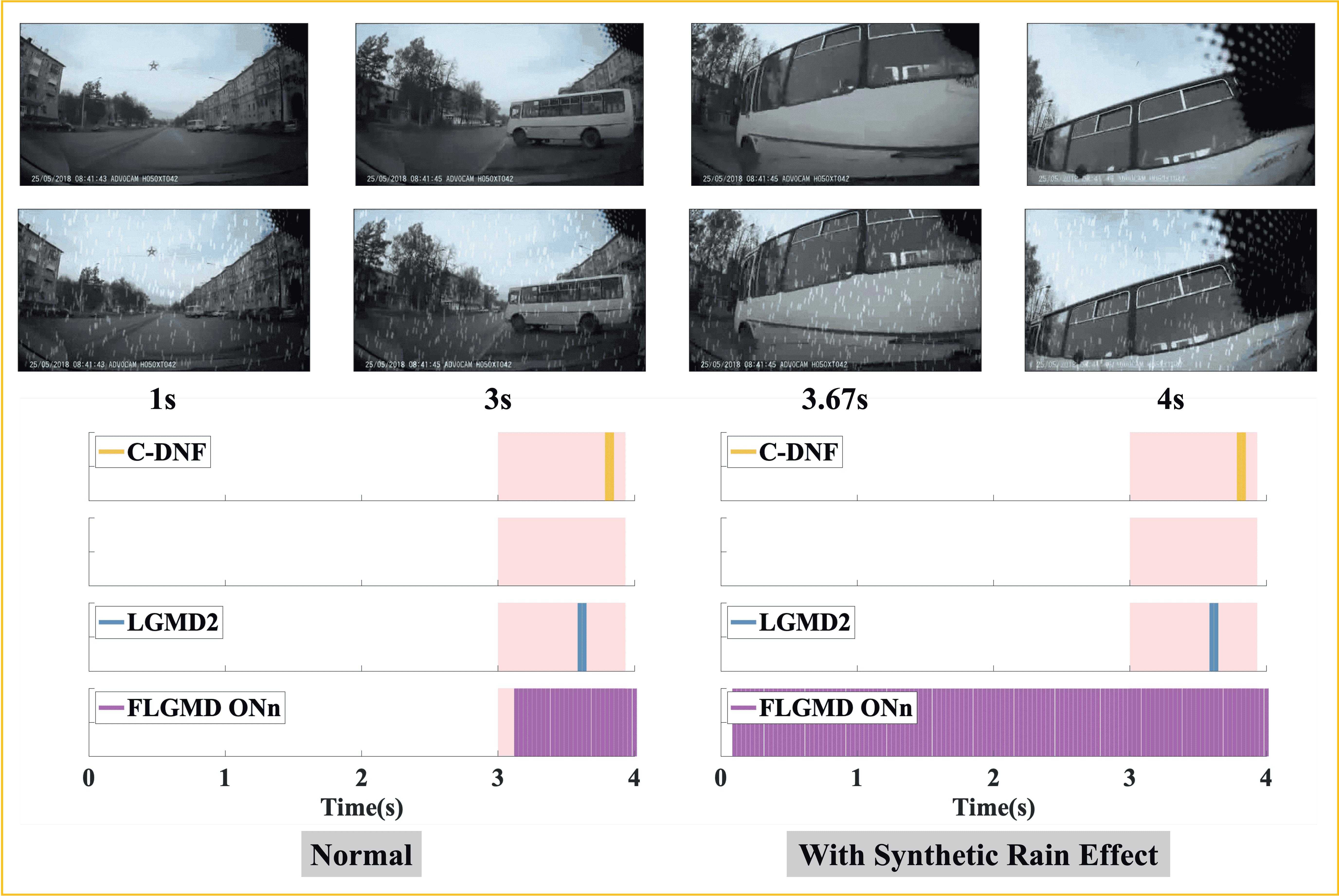} 
        \caption{ }
        \label{fig:c19} 
    \end{subfigure}
    \begin{subfigure}[t]{0.48\textwidth} 
        \centering
        \includegraphics[width=\textwidth]{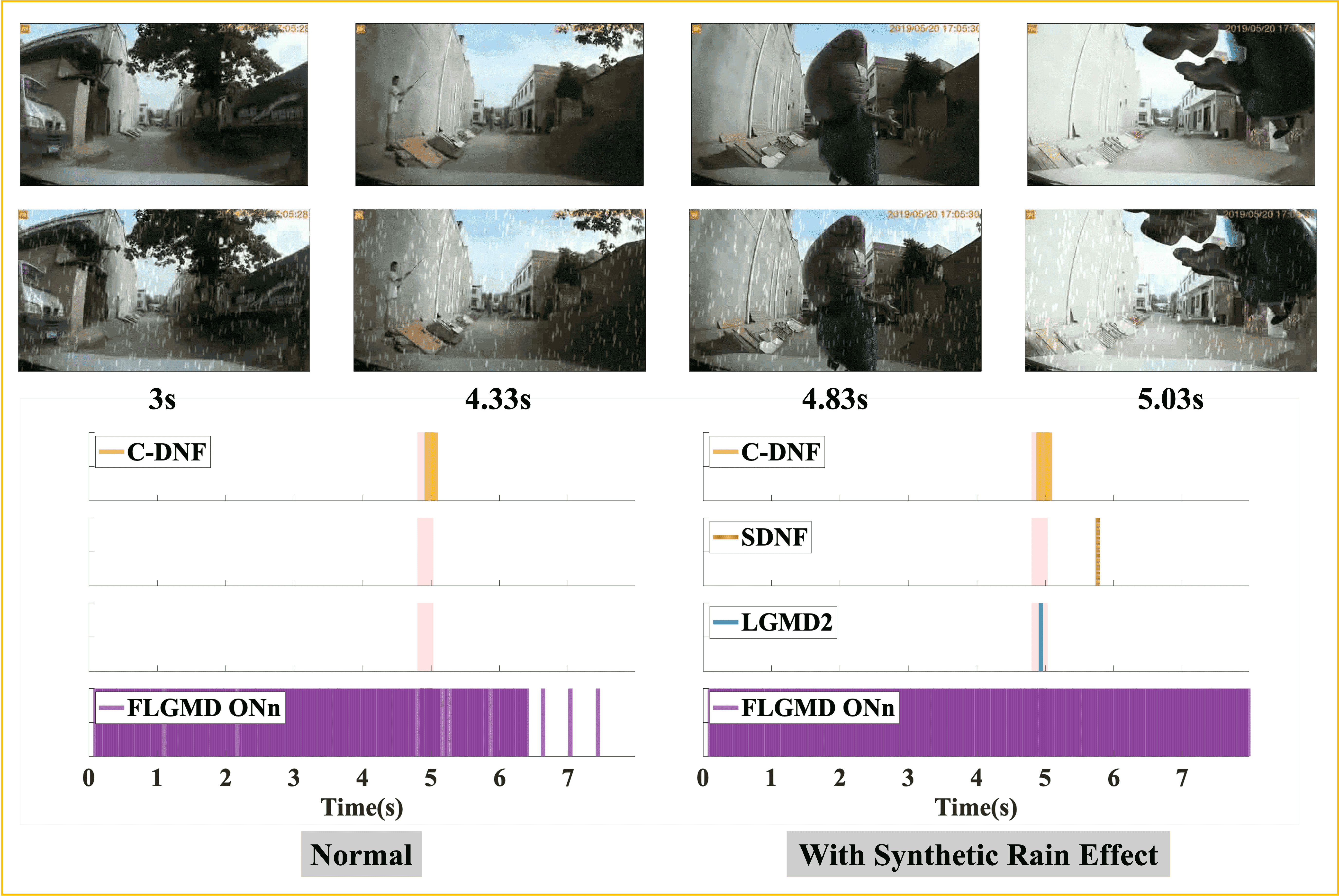} 
        \caption{ } 
        \label{fig:tc08}
    \end{subfigure}
    
    \caption{Comparative experiment results on recorded real-world stimuli with or without synthetic rain effect. The first row of the figures illustrates snapshots of the original tested real-world stimuli, while the second row depicts the same stimuli with synthetic rain effects applied. The manually identified collision period is highlighted with a pink shade in each sub-figure. (a) Comparative experiment results when tested with a stimulus depicting a bus collision. (b) Comparative experiment results when tested with a stimulus depicting a balloon hits the car. }
    \label{fig:realworld and rain}
%    \vspace{-10pt}
\end{figure*}

Interestingly, the LGMD2 model exhibited increased accuracy when tested on stimuli with synthetic rain effects. This phenomenon might be due to the rain effects acting as a form of signal augmentation, enhancing the model's ability to distinguish collision scenarios (see Fig.\ref{fig:tc08}). Conversely, the SDNF model showed a slight decrease in accuracy, likely because its binary pre-processing step misinterpreted raindrops as strong stimuli, leading to erroneous detections. The FLGMD ONn model produced timely collision alerts on certain real-world stimuli. However, when synthetic rain effects were applied, these alerts were advanced to the beginning of the stimulus, likely due to misinterpretation of the raindrops as motion cues.

Detailed accuracy metrics for the two tested datasets are presented in Table \ref{tab:exp result realworld}, while comparative experiment results for two sample stimuli are illustrated in Fig. \ref{fig:realworld and rain}. These results unequivocally demonstrate the robustness and effectiveness of the proposed C-DNF model, particularly in maintaining its accuracy and selectivity under challenging conditions, such as synthetic rain effects.

\begin{table}[t]
% \vspace{-10pt}
\caption{Accuracy of Various Models to Real-World Data}
\centering
\begin{tabular}{c c c c c}
\hline
Stimuli & C-DNF & SDNF & LGMD2 & F-LGMD ONn \\
\hline
\thead{Normal} & \textbf{77.14\%} & 22.86\% & 34.29\% & 28.57\% \\
\hline
\thead{With Synthetic\\ Rain effect} & \textbf{77.14\%} & 20\% & 40\% & 0\% \\
\hline
\end{tabular}
\label{tab:exp result realworld}
\end{table}

\section{Discussion and Conclusion}
\label{Sec:Dis}
This paper presents a DNF-based looming perception model integrated with ON/OFF visual contrast processing. Lateral excitation is implemented using a non-negative normalized Gaussian kernel within each contrast field. Compared to the previous single-field DNF-based method (SDNF) and two state of the arts (LGMD2 and FLGMD ONn), the proposed framework exhibits improved looming selectivity for incoherent stimuli. Its effectiveness is further validated through experiments with complex real-world scenarios, including those featuring synthetic rain effects, highlighting its effectiveness in challenging environments with inconsistent visual cues.

The proposed C-DNF maintains its robust looming selectivity and perception accuracy for two key reasons: first, the enhancement of signals in both ON and OFF contrasts through lateral excitation; second, the application of a DoG kernel and a fixed resting level in the Summation field, which effectively filters isolated noise. The lateral excitation scale was precisely set to include $8$ neighboring neurons. A larger scale, while further enhancing contrast signals, could overly activate neurons in the Summation field. However, as regulated by the DoG kernel, increased excitation would also result in stronger inhibition, potentially reducing the likelihood of generating timely collision alerts.

The LGMD2 model also retains its looming selectivity on synthetic incoherent stimuli, likely due to the spatial-temporal inhibition mechanisms at various scales within its structure, as no evidence of lateral excitation is present in its formulation. Notably, the LGMD2 model demonstrates even higher perception accuracy on real-world stimuli with synthetic rain effects. This improvement may result from the noise occasionally amplifying the input signal's intensity, thereby enhancing the model's response.

As a looming perception model, the proposed C-DNF exhibits responsiveness to receding stimuli, which, in some scenarios, could result in false collision alerts. While the dynamic threshold mechanism employed by the previous SDNF model effectively suppresses responses to receding stimuli, its application to the C-DNF reveals limitations, particularly when addressing translating stimuli. This highlights the necessity for novel strategies to further refine the looming selectivity of the C-DNF.

In general, the proposed C-DNF is the first attempt on formulating DNF-based looming perception model accounting for ON/OFF visual contrast operated in multiple fields, demonstrating robust performance under challenging conditions, including simulated raindrops. This suggests its potential to effectively handle artificial noise types, such as Gaussian or salt-and-pepper noise. Additionally, as multi-field DNF models are commonly utilized for more complex tasks like learning and memory simulation, the C-DNF model holds significant promise for integration with these frameworks. Such integration could enable the model to support more sophisticated cognitive functions, including navigation and decision-making, thereby extending its utility beyond fundamental perception tasks.

\section*{Acknowledgment}
This research has been supported by the National Natural Science Foundation of China under Grant No. 62376063. 
\textcolor{blue}{\textit{Corresponding author: Qinbing Fu} (qifu@gzhu.edu.cn)}

\bibliographystyle{IEEEtran}
\bibliography{mDNF_ref}

% \bibitem{b1} G. Eason, B. Noble, and I. N. Sneddon, ``On certain integrals of Lipschitz-Hankel type involving products of Bessel functions,'' Phil. Trans. Roy. Soc. London, vol. A247, pp. 529--551, April 1955.
% \bibitem{b2} J. Clerk Maxwell, A Treatise on Electricity and Magnetism, 3rd ed., vol. 2. Oxford: Clarendon, 1892, pp.68--73.
% \bibitem{b3} I. S. Jacobs and C. P. Bean, ``fine particles, thin films and exchange anisotropy,'' in Magnetism, vol. III, G. T. Rado and H. Suhl, Eds. New York: Academic, 1963, pp. 271--350.
% \bibitem{b4} K. Elissa, ``Title of paper if known,'' unpublished.
% \bibitem{b5} R. Nicole, ``Title of paper with only first word capitalized,'' J. Name Stand. Abbrev., in press.
% \bibitem{b6} Y. Yorozu, M. Hirano, K. Oka, and Y. Tagawa, ``Electron spectroscopy studies on magneto-optical media and plastic substrate interface,'' IEEE Transl. J. Magn. Japan, vol. 2, pp. 740--741, August 1987 [Digests 9th Annual Conf. Magnetics Japan, p. 301, 1982].
% \bibitem{b7} M. Young, The Technical Writer's Handbook. Mill Valley, CA: University Science, 1989.
% \end{thebibliography}
% \vspace{12pt}

\end{document}